\documentclass{article} 
\usepackage{iclr2027_conference,times}
\usepackage{CJKutf8}

\usepackage{amsmath,amsfonts,bm}

\def\eqref#1{equation~\ref{#1}}

\def\1{\bm{1}}

\DeclareMathAlphabet{\mathsfit}{\encodingdefault}{\sfdefault}{m}{sl}
\SetMathAlphabet{\mathsfit}{bold}{\encodingdefault}{\sfdefault}{bx}{n}

\usepackage{microtype}
\usepackage[hidelinks]{hyperref}
\usepackage{xurl}
\usepackage{algorithm}
\usepackage{algpseudocode}
\usepackage{graphicx}
\usepackage{subcaption}
\usepackage{amsmath,amssymb,amsthm}
\usepackage{booktabs}
\usepackage{placeins}

\newtheorem{assumption}{Assumption}
\newtheorem{theorem}{Theorem}
\newtheorem{lemma}{Lemma}
\newtheorem{corollary}{Corollary}
\newtheorem{proposition}{Proposition}

\title{Accelerating Muon with Spectral Residual Correction}

\author{
  Zhixin Ren$^{1}$, Yao Lyu$^{2}$\,$^{3}$,  Shengbo Eben Li$^{2}$\,$^{4}$\textdagger,Liping Zhang$^{1}$\textdagger  \\
  $^{1}$ Department of Mathmatical Science, Tsinghua University\\
  $^{2}$School of Vehicle and Mobility, Tsinghua University\\
  $^{3}$SunRisingAI Lab\,\,
  $^{4}$College of AI, Tsinghua University\\
  \textdagger Corresponding author
}

\iclrfinalcopy 
\begin{document}
\begin{CJK*}{UTF8}{gbsn}

\maketitle

\fancyhead{}
\renewcommand{\headrulewidth}{0pt}
\begin{abstract}
Muon constructs matrix updates by applying a matrix-sign transformation to momentum. A common route to accelerating Muon is to modify the gradient or momentum before this transformation. We explore a complementary direction: refining the parameter step while retaining Muon's momentum recursion. Derived from the parameter subproblem of the ADMM-Inspired Momentum framework, AIMuon uses the gradient--momentum residual to correct the tentative spectral step. The residual undergoes a separate matrix-sign transformation, while spectral--nuclear duality determines the relative weighting of the momentum and residual directions. In practice, both directions are approximated with Newton--Schulz iterations and their combination is rescaled to match Muon's update magnitude. We establish a stochastic stationarity bound for the corresponding exact iteration. On GPT-2-small, GPT-2-medium, and Qwen3-0.6B,
AIMuon reaches Muon's final validation loss with
8.98\%, 10.05\%, and 6.49\% fewer optimization steps,
respectively, while achieving lower final validation
loss than Muon, Muon-Nesterov, and MONA.
\end{abstract}

\section{Introduction}

Improving optimization efficiency is an important goal
in neural network training.
Muon~\citep{jordan2024muon} has emerged as an effective
optimizer for matrix parameters in large language
model training~\citep{liu2025muonscalablellmtraining,qwen2026design}.
Unlike coordinate-wise adaptive methods, Muon approximately
orthogonalizes the momentum matrix using Newton--Schulz
iterations~\citep{schulz1933iterative}, producing a
matrix-sign-like update that acts on its singular structure.
Its empirical performance motivates exploring how
spectral updates can be further accelerated.

An established approach to acceleration is to improve
how gradient information is accumulated into momentum.
Adan~\citep{xie2024adan} incorporates gradient differences
into both first- and second-order moment estimates through
a reformulation of Nesterov acceleration.
ALTO~\citep{zhao2025exploring} uses gradient differences
as a low-cost curvature proxy and applies their EMA
to correct the gradient entering the momentum estimates.
MONA~\citep{li2026mona} incorporates related
gradient-difference information into Muon's momentum
before applying the matrix-sign transformation.
These methods illustrate how modifying momentum construction
can improve optimization.

We explore a complementary perspective: accelerating
optimization through the construction of the parameter update.
The distinction is particularly relevant in spectral geometry.
For a shared fixed linear map, transforming a mixture
of momentum and residual is equivalent to transforming
the two signals separately and then combining them.
For the nonlinear matrix-sign map, these constructions
generally produce different update directions.
The treatment of gradient and momentum information
within the parameter update therefore offers a design
opportunity beyond modifying how momentum is accumulated.

The ADMM-Inspired Momentum (AIM)
framework~\citep{ren2026momentum} provides a starting point
for this approach.
AIM separates three interacting choices in optimizer
design: the penalty determines the update geometry,
the multiplier update or filtering rule determines
how gradient information is accumulated,
and the treatment of the objective-containing parameter
subproblem determines the parameter correction.
The study further highlights that more carefully
designed optimization iterations for the parameter
subproblem can lead to improved algorithms.

Since AIM also recovers Muon's direction under
spectral geometry, we extend this design principle
to propose \textbf{AIMuon}
(\textbf{A}DMM-\textbf{I}nspired \textbf{M}oment\textbf{u}m
\textbf{O}rthogonalized by \textbf{N}ewton--Schulz).
While retaining ordinary EMA momentum,
AIMuon constructs an additional correction by solving
a locally linearized parameter subproblem.
Its solution jointly determines the gradient--momentum
residual, its independent matrix-sign transformation,
and its nuclear-norm weight.
Combining the resulting residual direction with
the spectral momentum direction yields a new update
through parameter-subproblem design.

Our contributions are summarized as follows:
\begin{itemize}
    \item We derive \textbf{AIMuon} by applying AIM's
    parameter-update design perspective to spectral
    optimization.
    Solving a locally linearized parameter subproblem
    jointly yields the gradient--momentum residual,
    its independent spectral transformation,
    and its nuclear-norm weight,
    while retaining ordinary EMA momentum.

    \item We clarify how independent spectral transformations
    and nuclear-norm weighting shape the update direction.
    Separately transforming momentum and residual
    can produce a different direction from
    transforming their mixture.
    Moreover, RMS matching removes the nuclear-norm
    scale of a single direction but preserves
    the relative nuclear-norm weights that shape
    the combined momentum and residual direction.

    \item We establish a stochastic nonconvex stationarity
    bound for the unnormalized AIMuon iteration with exact
    matrix-sign transformations, characterizing
    the optimization error and the variance-dependent
    stochastic error.

    \item Experiments on GPT-2-small, GPT-2-medium, and
    Qwen3-0.6B show lower final validation loss than
    independently tuned Muon, Muon-Nesterov, and MONA,
    with consistent improvements across three seeds
    on GPT-2-small.
    Ablations demonstrate the effectiveness of
    independently orthogonalizing the residual
    and examine the contribution of nuclear-norm weighting.
    Sensitivity and batch-size experiments further
    characterize the effects of correction strength
    and batch size.
\end{itemize}

\section{Background and Related Work}
\label{sec:preliminaries}

We review related optimizer designs, introduce
the AIM framework, and explain the spectral geometry
underlying AIMuon.

\subsection{Related Work}
\label{sec:related_work}

\paragraph{Momentum and adaptivity.}
Momentum and adaptivity are central components of modern
optimization algorithms.
Classical Polyak momentum~\citep{polyak1964some}
accumulates past gradients, while Nesterov
acceleration~\citep{nesterov1983method} modifies how
momentum information contributes to the update.
Adaptive methods further rescale updates using gradient
statistics.
Adam~\citep{kingma2015adam} combines first- and
second-moment estimates,
NAdam~\citep{dozat2016incorporating} incorporates
Nesterov momentum into this adaptive framework,
and AdamW~\citep{loshchilov2019decoupled} decouples
weight decay from adaptive scaling.
More recent methods such as RAD~\citep{lyu2025conformal}
derive adaptive updates from alternative dynamical
principles.
Building on RAD's adaptive preconditioner within the
AIM framework, RADAR~\citep{ren2026momentum} modifies
both momentum construction and the parameter-update
subproblem.
AIMuon retains ordinary EMA momentum and develops
the parameter-subproblem design route in spectral geometry.

\paragraph{Structured preconditioning and normalization.}
Beyond coordinate-wise adaptivity, many optimizers exploit
structure within matrix or tensor parameters.
Adafactor~\citep{shazeer2018adafactor} approximates
second-moment estimates using row- and column-wise
factors, reducing optimizer memory.
Adam-mini~\citep{zhang2025adammini} replaces Adam's
element-wise adaptive learning rates with shared learning
rates over selected parameter blocks.
Shampoo~\citep{gupta2018shampoo} builds Kronecker-factored
second-order preconditioners along the dimensions of
tensor parameters, while SOAP~\citep{vyas2025soap}
applies Adam-style adaptivity in a rotated basis induced
by Shampoo-style preconditioners.
Matrix-geometric methods such as spectral
descent~\citep{carlson2015spectral}, Muon, and
PolarGrad~\citep{lau2025polargrad} exploit spectral or
polar structure to construct matrix-aware update directions,
with PolarGrad further introducing nuclear-norm-based scaling.
NorMuon~\citep{li2025normuon} augments Muon with neuron-wise
adaptive normalization based on second-moment statistics,
while AdaMuon~\citep{si2025adamuon} combines element-wise
adaptivity with orthogonalized matrix updates.

\paragraph{Scalable orthogonalized optimizers.}
Recent work improves the efficiency and scalability
of orthogonalized matrix updates.
\citet{liu2025muonscalablellmtraining} study the scaling,
update normalization, and distributed implementation
of Muon for large-scale language-model training.
Dion~\citep{ahn2025dion} reduces computation and communication
through low-rank orthonormalized updates with error feedback.
ROOT~\citep{he2025root} adapts its orthogonalization
coefficients to matrix dimensions and uses a proximal
formulation to improve robustness to outlier noise.
These developments extend orthogonalized optimization
to practical large-scale training.

\subsection{AIM Framework}
\label{sec:prelim_muon}

Consider the stochastic optimization problem
\begin{equation}
    \min_{\Theta}\;
    F(\Theta)
    :=
    \mathbb{E}_{\xi}\!\left[f(\Theta;\xi)\right],
    \label{eq:prelim_problem}
\end{equation}
with stochastic gradient
$G_k=\nabla f(\Theta_k;\xi_k)$.
For compatible matrices, we use
$\langle A,B\rangle=\operatorname{tr}(A^\top B)$
for the Frobenius inner product and denote the
Frobenius, spectral, and nuclear norms by
$\|\cdot\|_F$, $\|\cdot\|_2$, and $\|\cdot\|_*$,
respectively.

The ADMM-Inspired Momentum (AIM)
framework~\citep{ren2026momentum} introduces an auxiliary
copy $Y$ of the parameters $\Theta$, with the constraint
$Y=\Theta$.
It separates optimizer construction into an
auxiliary-variable subproblem, an objective-containing
parameter subproblem, and a multiplier update:
\begin{equation}
\begin{aligned}
Y_{k+1}
&\in
\arg\min_Y
\left\{
    \langle M_k,Y-\Theta_k\rangle
    +
    \frac{\rho_{1,k}}{2}\psi(Y-\Theta_k)
\right\},
\\
\Theta_{k+1}
&\in
\arg\min_\Theta
\left\{
    F(\Theta)
    +
    \langle M_k,Y_{k+1}-\Theta\rangle
    +
    \frac{\rho_{2,k}}{2}\psi(Y_{k+1}-\Theta)
\right\},
\\
M_{k+1}
&=
M_k
+
(1-\beta)\frac{\rho_{2,k}}{2}D_{k+1}.
\end{aligned}
\label{eq:prelim_aim_updates}
\end{equation}
Here, $M_k$ is the multiplier associated with $Y=\Theta$,
$\psi$ is a convex penalty function,
$\rho_{1,k},\rho_{2,k}>0$ are penalty parameters,
and $\beta\in[0,1)$.
The subgradient
$D_{k+1}\in\partial\psi(Y_{k+1}-\Theta_{k+1})$
is selected through the parameter-subproblem
optimality condition.

The auxiliary-variable subproblem constructs a tentative
point using the multiplier and the chosen penalty geometry.
The parameter subproblem incorporates the objective $F$
while penalizing displacement from this tentative point.
The multiplier update accumulates the discrepancy
between the two variables in the chosen geometry.
Combining this update with the optimality condition
of the $\Theta$-subproblem recovers Polyak momentum
in its EMA form~\citep{ren2026momentum}.

AIM thus separates three interacting principles
for optimizer design:
\emph{the penalty determines the update geometry};
\emph{the multiplier update or filtering rule determines
how gradient information is accumulated};
and \emph{the approximation of the objective-containing
parameter subproblem determines the form of
the parameter correction}.
These principles connect the geometry, momentum,
and parameter update to specific choices within
the framework.

Under Euclidean geometry, retaining the EMA recursion
and directly accepting the tentative point gives
SGD with Polyak momentum.
Approximating the $\Theta$-subproblem by linearizing
$F$ at the current iterate instead introduces
a correction proportional to $G_k-M_k$,
yielding a Nesterov-style update for suitable
penalty parameters~\citep{ren2026momentum}.
This correspondence illustrates how different
treatments of the parameter subproblem produce
different updates while retaining the same
momentum recursion.
It motivates exploring whether the same linearization
of $F$ in the $\Theta$-subproblem can yield a correction
that accelerates Muon under spectral-norm geometry.

\subsection{Spectral Geometry and the Muon Direction}
\label{sec:prelim_matrix_geometry}

To extend the preceding construction to Muon,
we first specify the spectral penalty in AIM
and establish its connection to Muon's update direction.

\paragraph{Why spectral geometry?}
\citet{yang2023spectral} study spectral conditions
for feature learning and connect spectral scaling
to maximal update parameterization ($\mu$P).
Their analysis motivates using spectral geometry
to characterize how matrix updates affect layer features.
To illustrate this connection, consider a linear layer
$h=Wx$, where
$W\in\mathbb{R}^{d_{\mathrm{out}}\times d_{\mathrm{in}}}$.
At fixed input, a weight update $\Delta W$ produces
$\Delta h=\Delta W x$, with
\begin{equation}
    \operatorname{RMS}(\Delta h)
    \leq
    \sqrt{\frac{d_{\mathrm{in}}}{d_{\mathrm{out}}}}
    \|\Delta W\|_2\operatorname{RMS}(x).
    \label{eq:spectral_feature_change}
\end{equation}
The dimension-adjusted spectral norm measures
the maximum amplification from input features
to output changes.
This feature-based interpretation motivates our choice
of spectral geometry for matrix updates.

\paragraph{Muon's constrained spectral update.}
Spectral geometry also preserves Muon's underlying
direction construction.
Let $A=U_r\Sigma_rV_r^\top$ be the compact SVD of
a nonzero matrix $A$, and define its matrix-sign,
or polar, direction as
\begin{equation}
    \operatorname{MSign}(A)
    :=
    U_rV_r^\top,
    \label{eq:prelim_msign}
\end{equation}
with $\operatorname{MSign}(0)=0$.
By spectral--nuclear duality,
\begin{equation}
    \|A\|_*
    =
    \max_{\|Z\|_2\leq1}\langle A,Z\rangle,
    \qquad
    \langle A,\operatorname{MSign}(A)\rangle
    =
    \|A\|_*.
    \label{eq:prelim_spectral_nuclear_duality}
\end{equation}
Consequently, for a momentum matrix $M_k$,
the constrained linear problem
\begin{equation}
    \min_{\Delta}\;
    \langle M_k,\Delta\rangle
    \quad\text{s.t.}\quad
    \|\Delta\|_2\leq\eta_k
    \label{eq:prelim_muon_constrained}
\end{equation}
admits the canonical solution
$-\eta_k\operatorname{MSign}(M_k)$.
This is Muon's basic spectral direction:
the constraint specifies the update geometry,
while its radius determines the update scale.

\paragraph{Spectral penalties in AIM.}
Muon's formulation above expresses spectral geometry
through a constraint, whereas AIM encodes update
geometry through the displacement penalty $\psi$.
To connect the two formulations, we consider
a penalized representation of the same geometry.
A familiar example is AdamW's preconditioned update,
which, ignoring decoupled weight decay, can be written as
\begin{equation}
    \arg\min_{\Delta}
    \left\{
        \langle m_k,\Delta\rangle
        +
        \frac{1}{2\eta_k}\|\Delta\|_{V_k}^2
    \right\}
    =
    -\eta_kV_k^{-1}m_k.
    \label{eq:prelim_preconditioned_penalty}
\end{equation}
Here, $\Delta$ and $m_k$ are vectors,
$m_k$ is the bias-corrected first moment,
$V_k$ is the positive-definite diagonal denominator
matrix, and
$\|\Delta\|_{V_k}^2=\langle\Delta,V_k\Delta\rangle$.
The quadratic displacement penalty determines
the preconditioned update geometry.

To express Muon's spectral geometry within AIM,
we instead choose
$\psi(X)=\|X\|_2^2$.
For a matrix-valued driving signal $A$, the resulting
generic subproblem is
\begin{equation}
    \min_{\Delta}
    \left\{
        \langle A,\Delta\rangle
        +
        \frac{\rho}{2}\|\Delta\|_2^2
    \right\},
    \qquad
    \rho>0.
    \label{eq:prelim_msign_penalty}
\end{equation}
By spectral--nuclear duality, choosing the canonical
polar solution gives
\begin{equation}
    \Delta^\star
    =
    -\frac{\|A\|_*}{\rho}\operatorname{MSign}(A).
    \label{eq:prelim_msign_penalty_solution}
\end{equation}
Appendix~\ref{app:msign_solution_set} characterizes
the full solution set.
The squared spectral-norm penalty therefore preserves
the matrix-sign direction while determining an
input-dependent radial magnitude through the
nuclear norm.

\paragraph{Recovering the Muon direction.}
With $\psi(X)=\|X\|_2^2$ and
$\Delta=Y-\Theta_k$, AIM's auxiliary-variable
subproblem becomes
\begin{equation}
    \min_{\Delta}
    \left\{
        \langle M_k,\Delta\rangle
        +
        \frac{\rho_{1,k}}{2}\|\Delta\|_2^2
    \right\}.
    \label{eq:prelim_spectral_subproblem}
\end{equation}
Applying Eq.~(\ref{eq:prelim_msign_penalty_solution})
with $A=M_k$ and $\rho=\rho_{1,k}$ gives
\begin{equation}
    Y_{k+1}
    =
    \Theta_k
    -
    \frac{\|M_k\|_*}{\rho_{1,k}}
    \operatorname{MSign}(M_k).
    \label{eq:prelim_spectral_y}
\end{equation}
Under the explicit EMA construction, directly assigning
$\Theta_{k+1}=Y_{k+1}$ and setting $\rho_{1,k}=1/\eta_k$
yields
\begin{equation}
    \Theta_{k+1}
    =
    \Theta_k
    -
    \eta_k\|M_k\|_*\operatorname{MSign}(M_k).
    \label{eq:prelim_spectral_update}
\end{equation}
This update has the nuclear-norm-weighted polar
structure of PolarGrad~\citep{lau2025polargrad},
with momentum as its input.
When its RMS magnitude is matched to Muon's,
the nuclear-norm factor $\|M_k\|_*$ cancels exactly.
Thus, with the same momentum input and RMS scaling,
this PolarGrad-form update is equivalent to Muon.

\section{AIMuon: Spectral Residual Correction}
\label{sec:aimuon}

Inspired by the Euclidean correspondence between
different treatments of the $\Theta$-subproblem
and SGD with momentum or Nesterov acceleration,
we construct AIMuon by changing how the
$\Theta$-subproblem is solved under spectral geometry.
We solve a local
surrogate obtained by linearizing the objective.
This construction jointly yields the gradient--momentum
residual, its independent spectral transformation,
and its nuclear-norm weight.

\subsection{Residual Correction from the AIM Parameter Subproblem}
\label{sec:aimuon_derivation}

We reindex the iteration so that the momentum update
comes first, followed by the spectral auxiliary-variable
update:
\begin{subequations}
\begin{align}
    M_k
    &=
    \beta M_{k-1}+(1-\beta)G_k,
    \label{eq:aimuon_momentum}
    \\
    Y_{k+1}
    &=
    \Theta_k
    -
    \frac{\|M_k\|_*}{\rho_{1,k}}
    \operatorname{MSign}(M_k),
    \label{eq:aimuon_y_solution}
\end{align}
\end{subequations}
where $M_{-1}=0$ and $G_k$ is the current
stochastic gradient.

Following the Euclidean construction, we derive
a correction to $Y_{k+1}$ from the $\Theta$-subproblem
\begin{equation}
    \min_{\Theta}
    \left\{
        F(\Theta)
        +
        \langle M_k,Y_{k+1}-\Theta\rangle
        +
        \frac{\rho_{2,k}}{2}
        \|\Theta-Y_{k+1}\|_2^2
    \right\}.
    \label{eq:aimuon_theta_subproblem}
\end{equation}
Linearizing $F$ at $\Theta_k$ using the current
stochastic gradient $G_k$ and defining
$E_k:=G_k-M_k$ gives
\begin{equation}
    \min_{\Theta}
    \left\{
        \langle E_k,\Theta-Y_{k+1}\rangle
        +
        \frac{\rho_{2,k}}{2}
        \|\Theta-Y_{k+1}\|_2^2
    \right\}.
    \label{eq:aimuon_theta_local}
\end{equation}
Here, $E_k$ is the gradient--momentum residual,
the difference between the current stochastic
gradient and the updated EMA momentum.
Applying Eq.~(\ref{eq:prelim_msign_penalty_solution})
yields
\begin{equation}
    \Theta_{k+1}
    =
    Y_{k+1}
    -
    \frac{\|E_k\|_*}{\rho_{2,k}}
    \operatorname{MSign}(E_k).
    \label{eq:aimuon_theta_refinement}
\end{equation}
The residual signal, its independent matrix-sign
transformation, and its nuclear-norm weight thus
follow jointly from the local parameter subproblem,
without an additional gradient evaluation.

Substituting Eq.~(\ref{eq:aimuon_y_solution})
and setting $\eta_k=1/\rho_{1,k}$ and
$c=\rho_{1,k}/\rho_{2,k}$ gives the unnormalized
AIMuon update with exact matrix-sign transformations:
\begin{equation}
    \Theta_{k+1}
    =
    \Theta_k
    -
    \eta_k
    \left[
        \|M_k\|_*\operatorname{MSign}(M_k)
        +
        c\,\|E_k\|_*\operatorname{MSign}(E_k)
    \right].
    \label{eq:aimuon_unscaled_update}
\end{equation}
The auxiliary-variable and parameter subproblems
produce separate spectral directions for the momentum
and residual, respectively.
Their nuclear norms and the ratio of penalty parameters
determine how these directions are combined.

\subsection{Practical Implementation}
\label{sec:aimuon_update}

\begin{algorithm}[t]
\caption{AIMuon}
\label{alg:aimuon}
\begin{algorithmic}[1]
\Require $\Theta_0$, $M_{-1}=0$, learning rates
$\{\eta_k\}_{k=0}^{T-1}$, momentum coefficient $\beta$,
correction coefficient $c$, weight decay $\lambda_{wd}$, numerical stabilizer $\varepsilon>0$
\For{$k=0,\ldots,T-1$}
    \State $G_k \gets \nabla f(\Theta_k;\xi_k)$
    \State $M_k \gets \beta M_{k-1}+(1-\beta)G_k$
    \State $E_k \gets G_k-M_k$
    \State $D_k^M \gets \operatorname{NS}_5(M_k)$
    \State $D_k^E \gets \operatorname{NS}_5(E_k)$
    \State $\widehat{\nu}_k^M
    \gets |\langle M_k,D_k^M\rangle|$
    \State $\widehat{\nu}_k^E
    \gets |\langle E_k,D_k^E\rangle|$
    \State $\lambda_k
    \gets
    c\frac{\widehat{\nu}_k^E}
    {\widehat{\nu}_k^M+\varepsilon}$
    \State $\widetilde{O}_k
    \gets
    \widehat{\nu}_k^M
    \left(D_k^M+\lambda_kD_k^E\right)$
    \State $O_k
    \gets
    0.2\sqrt{\max(m,n)}
    \frac{\|D_k^M\|_F}
    {\|\widetilde{O}_k\|_F+\varepsilon}
    \widetilde{O}_k$
    \State $\Theta_{k+1}
    \gets
    (1-\eta_k\lambda_{wd})\Theta_k-\eta_kO_k$
\EndFor
\end{algorithmic}
\end{algorithm}

Algorithm~\ref{alg:aimuon} summarizes the practical update
for a parameter matrix
$\Theta_k\in\mathbb{R}^{m\times n}$.
Here $\varepsilon>0$ stabilizes the ratios and
$\lambda_{wd}$ denotes weight decay.

\paragraph{Approximating the polar directions.}
We independently approximate the momentum and residual
matrix-sign directions using five Newton--Schulz iterations:
\begin{equation}
    D_k^M
    =
    \operatorname{NS}_5(M_k),
    \qquad
    D_k^E
    =
    \operatorname{NS}_5(E_k).
    \label{eq:aimuon_ns_directions}
\end{equation}

\paragraph{Matrix-wise nuclear-norm weighting.}
Using Eq.~(\ref{eq:prelim_spectral_nuclear_duality}),
we estimate each nuclear norm by replacing the exact
polar factor with its NS5 approximation:
\begin{equation}
    \widehat{\nu}_k^M
    =
    \left|\langle M_k,D_k^M\rangle\right|,
    \qquad
    \widehat{\nu}_k^E
    =
    \left|\langle E_k,D_k^E\rangle\right|.
    \label{eq:aimuon_nuclear_estimates}
\end{equation}
These estimates reuse the computed spectral directions
and avoid an additional singular-value decomposition.
For each parameter matrix, the relative residual weight is
\begin{equation}
    \lambda_k
    =
    c\frac{\widehat{\nu}_k^E}
    {\widehat{\nu}_k^M+\varepsilon}.
    \label{eq:aimuon_lambda}
\end{equation}
The combined direction is
\[
    \widetilde{O}_k
    =
    \widehat{\nu}_k^M
    \left(D_k^M+\lambda_kD_k^E\right).
\]
The residual-to-momentum nuclear-norm ratio adjusts
the relative contribution of the two directions
across matrices and throughout training.

\paragraph{RMS matching.}
The additional residual correction changes both
the direction and magnitude of the update.
To isolate the benefit of the directional change,
we match the Frobenius norm of the combined update
to that of Muon constructed from the same momentum.
Following Muon's scaling
convention~\citep{liu2025muonscalablellmtraining},
we set
\begin{equation}
    O_k
    =
    0.2\sqrt{\max(m,n)}
    \frac{\|D_k^M\|_F}
    {\|\widetilde{O}_k\|_F+\varepsilon}
    \widetilde{O}_k.
    \label{eq:aimuon_final_direction}
\end{equation}
The reference update is
$0.2\sqrt{\max(m,n)}D_k^M$.
This matching controls the magnitude change introduced
by the residual term, allowing us to assess
the benefit of the combined spectral direction
at the same update scale.

\subsection{Conceptual Insights}
\label{sec:aimuon_interpretation}

\paragraph{Separate spectral transformations yield distinct directions.}
Under Euclidean geometry or a shared fixed linear
preconditioning map, combining signals before
transformation is equivalent to combining their
separately transformed outputs.
This equivalence generally fails in spectral geometry.
Writing
$\mathcal{D}(A)=\|A\|_*\operatorname{MSign}(A)$,
we generally have
\begin{equation}
    \mathcal{D}(M_k)+c\mathcal{D}(E_k)
    \neq
    \mathcal{D}(M_k+cE_k).
    \label{eq:aimuon_separate_transform}
\end{equation}
The two sides can be non-collinear, so RMS matching
does not remove their directional difference.
Thus, the correction derived from the parameter
subproblem generally cannot be absorbed into
the same momentum--residual mixture before
the spectral transformation.

\paragraph{Nuclear norms shape relative contributions after normalization.}
For the single momentum direction, RMS matching
removes the nuclear-norm factor, connecting
the PolarGrad-form update to Muon.
For AIMuon's combined direction
$U_k=\mathcal{D}(M_k)+c\mathcal{D}(E_k)$,
factoring out the common scale $\|M_k\|_*$
when $M_k\neq0$ gives
\begin{equation}
    \frac{U_k}{\|M_k\|_*}
    =
    \operatorname{MSign}(M_k)
    +
    c\frac{\|E_k\|_*}{\|M_k\|_*}
    \operatorname{MSign}(E_k).
    \label{eq:aimuon_relative_direction}
\end{equation}
RMS matching removes the common scale but preserves
the relative nuclear-norm weight.
The same spectral penalty therefore plays different
roles in the two constructions: its nuclear-norm
factor changes only the magnitude of a single
direction, but shapes the combined direction
when momentum and residual are transformed separately.

\paragraph{Component orthogonalization versus final-update structure.}
Muon uses an approximately orthogonalized momentum
matrix as its update direction.
AIMuon instead combines two separately approximately
orthogonalized matrices.
Their weighted sum generally no longer has a polar
structure, and RMS matching only rescales it.
This distinction motivates a further question:
how much of the benefit of spectral optimizers
comes from the orthogonal structure of the final
update, and how much comes from transforming
and combining the spectral structures of its components?
AIMuon provides a concrete construction for
investigating this question.

\subsection{Stochastic Convergence of the Exact Update}

\label{sec:aimuon_convergence}
We analyze the unnormalized iteration
in \eqref{eq:aimuon_unscaled_update} with exact
matrix-sign transformations.
The following result summarizes the constant-stepsize,
constant-batch case.

\begin{corollary}[Constant-stepsize stationarity bound]
\label{cor:aimuon_constant_stepsize}
Suppose Assumptions~\ref{ass:aimuon_smooth_fro}
and~\ref{ass:aimuon_noise_fro} hold: $F$ is lower bounded
and $L_F$-smooth in the Frobenius norm, and the stochastic
gradients are conditionally unbiased with conditional
variance bounded by $\sigma_F^2/B$.
Let $M_{-1}=0$, $\Theta_0$ be deterministic,
$0\leq\beta<1$, and $0<c<4$. Define
\[
r=\min(m,n),\qquad
\kappa_c=\frac{4-c}{4(1+2c)},\qquad
\Delta_F=F(\Theta_0)-F_{\inf}.
\]
For a constant stepsize satisfying
$0<\eta\leq\kappa_c/(rL_F)$, the iterates satisfy
\begin{equation}
\frac{1}{T}\sum_{k=0}^{T-1}
\mathbb{E}\|\nabla F(\Theta_k)\|_F^2
\leq
\frac{4\Delta_F}{c\kappa_c\eta T}
+\frac{4r\sigma_F^2}{c\kappa_c^2B}.
\label{eq:aimuon_main_stationarity}
\end{equation}
\end{corollary}

The first term decreases with the number of iterations,
while the second captures the contribution of stochastic
gradient noise and decreases with batch size.
The general result for nonincreasing stepsizes and
varying batch sizes, together with its proof, appears
in Appendix~\ref{app:aimuon_stochastic_fro}.

\FloatBarrier
\section{Experiments}
\label{sec:experiments}

\subsection{Experimental Setup}
\label{sec:pretraining_setup}

We pretrain GPT-2-small on FineWeb-Edu and
GPT-2-medium and Qwen3-0.6B on
FineWeb~\citep{penedo2024fineweb}, using approximately
2.48B, 6.9B, and 12B tokens, respectively.
All models are trained from scratch on eight
NVIDIA H100 GPUs.

We compare AIMuon with Muon (Polyak momentum),
Muon-Nesterov, and MONA.
Each optimizer receives independent sequential
learning-rate and momentum tuning based on terminal
validation loss.
Within each setting and seed, methods share
initialization, training-data order, and validation data.
GPT-2-small results are reported across seeds
42, 43, and 44; GPT-2-medium and Qwen3-0.6B use seed 42.
Full training configurations, search budgets, and
selected hyperparameters are provided in
Appendix~\ref{app:experiment_details}.

\subsection{Pretraining Performance}
\label{sec:main_results}

\begin{table}[t]
    \centering
    \caption{
    Final validation loss and perplexity.
    GPT-2-small results are mean $\pm$ standard
    deviation across three seeds.
    GPT-2-medium and Qwen3-0.6B use seed 42.
    }
    \label{tab:main_results}
    \small
    \setlength{\tabcolsep}{3pt}
    \begin{tabular}{lcccccc}
        \toprule
        & \multicolumn{2}{c}{GPT-2-small}
        & \multicolumn{2}{c}{GPT-2-medium}
        & \multicolumn{2}{c}{Qwen3-0.6B} \\
        \cmidrule(lr){2-3}
        \cmidrule(lr){4-5}
        \cmidrule(lr){6-7}
        Method
        & Loss $\downarrow$ & PPL $\downarrow$
        & Loss $\downarrow$ & PPL $\downarrow$
        & Loss $\downarrow$ & PPL $\downarrow$ \\
        \midrule
        Muon
        & $3.1980 \pm 0.0005$
        & $24.4825 \pm 0.0122$
        & 3.0255 & 20.6053
        & 2.7859 & 16.2136 \\
        Muon-Nesterov
        & $3.1996 \pm 0.0005$
        & $24.5218 \pm 0.0129$
        & 3.0195 & 20.4806
        & 2.7830 & 16.1671 \\
        MONA
        & $3.2210 \pm 0.0010$
        & $25.0536 \pm 0.0258$
        & 3.0316 & 20.7295
        & 2.7824 & 16.1573 \\
        AIMuon
        & $\mathbf{3.1945 \pm 0.0005}$
        & $\mathbf{24.3983 \pm 0.0134}$
        & \textbf{3.0187} & \textbf{20.4637}
        & \textbf{2.7819} & \textbf{16.1501} \\
        \bottomrule
    \end{tabular}
\end{table}

\begin{figure}[t]
    \centering
    \begin{subfigure}[t]{0.32\linewidth}
        \centering
        \includegraphics[width=\linewidth]
        {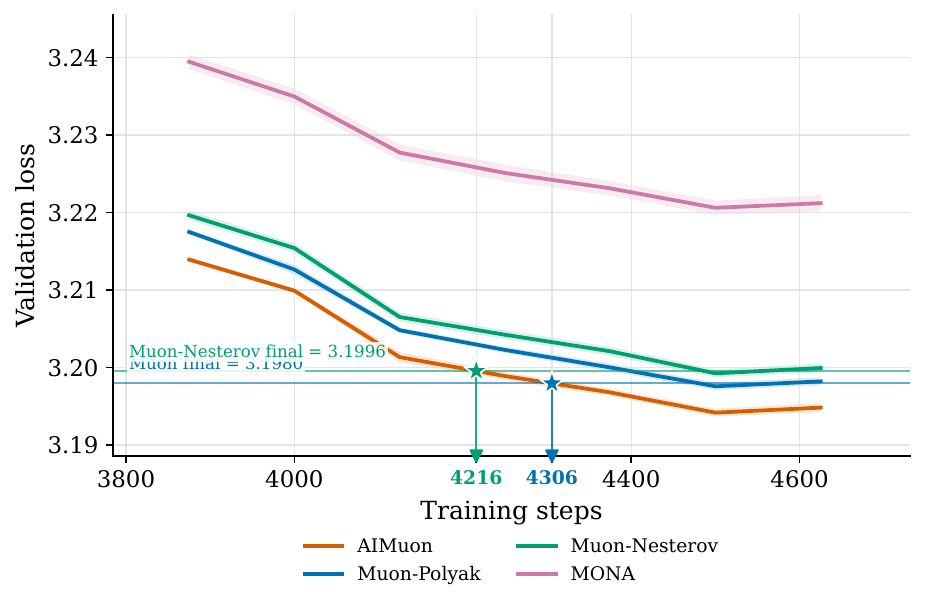}
        \caption{GPT-2-small.}
        \label{fig:small_pretraining_valid}
    \end{subfigure}
    \hfill
    \begin{subfigure}[t]{0.32\linewidth}
        \centering
        \includegraphics[width=\linewidth]
        {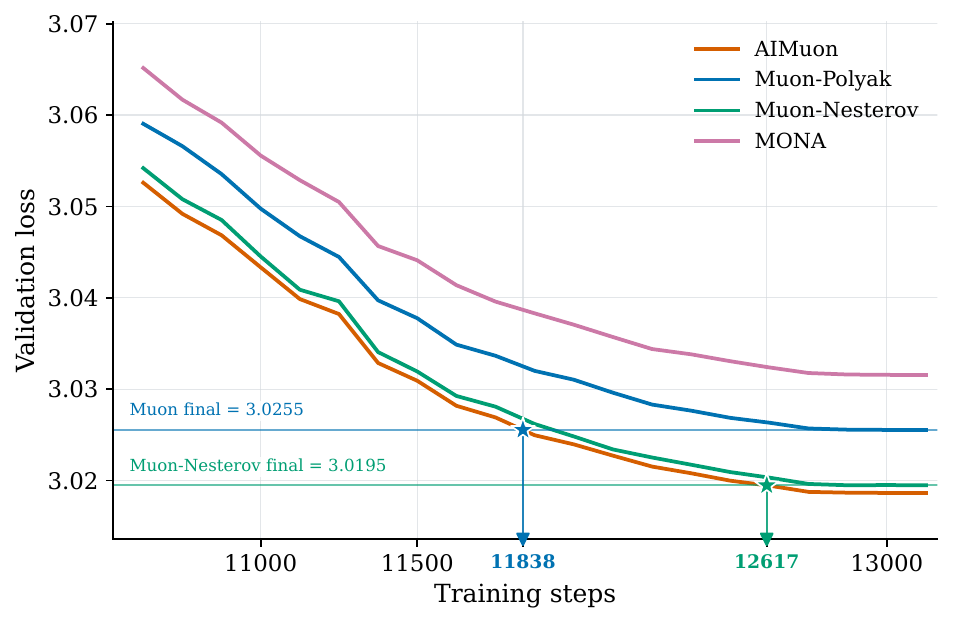}
        \caption{GPT-2-medium.}
        \label{fig:medium_pretraining_valid}
    \end{subfigure}
    \hfill
    \begin{subfigure}[t]{0.32\linewidth}
        \centering
        \includegraphics[width=\linewidth]
        {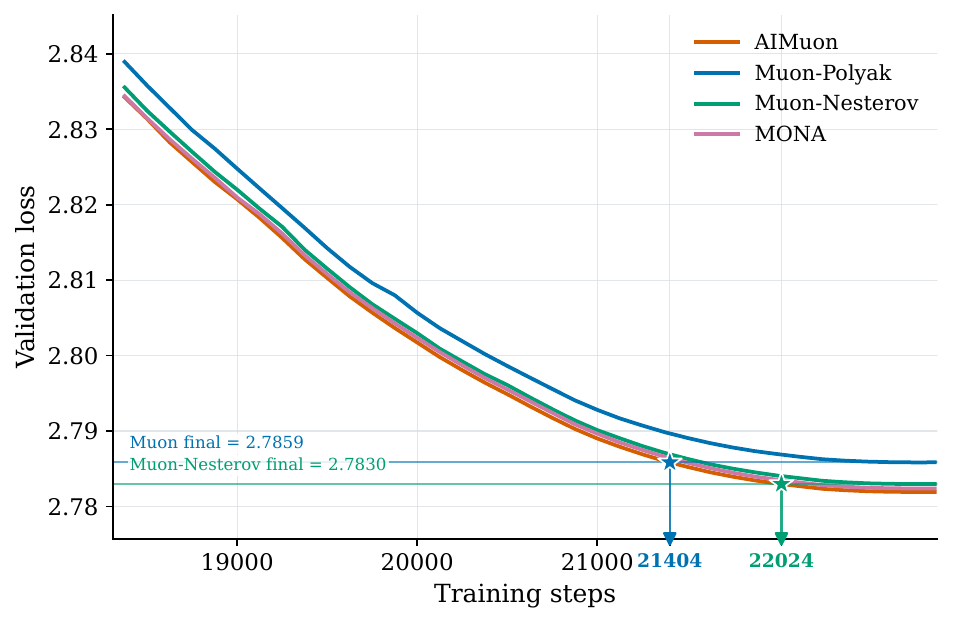}
        \caption{Qwen3-0.6B.}
        \label{fig:qwen_pretraining_valid}
    \end{subfigure}
    \caption{
    Validation loss versus optimization steps
    during the final portion of pretraining.
    GPT-2-small curves show the mean validation loss
    across three seeds; GPT-2-medium and Qwen3-0.6B
    use seed 42.
    Each optimizer uses independently selected
    learning-rate and momentum settings.
    }
    \label{fig:pretraining_validation}
\end{figure}

Table~\ref{tab:main_results} reports final validation
loss and perplexity, while
Figure~\ref{fig:pretraining_validation} shows
the late-training validation trajectories.
AIMuon achieves the lowest final validation loss
and perplexity across all three settings,
improving both the final pretraining performance
and the number of optimization steps required
to reach the baseline losses.

\paragraph{Improvements over Muon.}
AIMuon reduces Muon's final validation loss
from 3.1980 to 3.1945 on GPT-2-small,
from 3.0255 to 3.0187 on GPT-2-medium,
and from 2.7859 to 2.7819 on Qwen3-0.6B.
The corresponding perplexities decrease from
24.4825 to 24.3983, from 20.6053 to 20.4637,
and from 16.2136 to 16.1501, respectively.
On GPT-2-small, AIMuon outperforms every baseline
in each of the three matched seeds,
with paired loss reductions over Muon ranging
from approximately 0.0024 to 0.0041.
These results support the effectiveness of adding
a separately orthogonalized residual correction
to Muon's momentum-based spectral update.

\paragraph{Comparison with Muon-Nesterov and MONA.}
AIMuon also improves upon methods that modify
the signal entering the spectral transformation.
Compared with Muon-Nesterov, AIMuon lowers
the final validation loss by 0.0051 on GPT-2-small,
0.0008 on GPT-2-medium, and 0.0011 on Qwen3-0.6B.
On GPT-2-medium and Qwen3-0.6B, where Nesterov
already improves upon Polyak Muon,
AIMuon achieves a further reduction.
On GPT-2-small, AIMuon improves upon both variants,
whereas Muon-Nesterov finishes above Polyak Muon.
AIMuon also attains lower final loss than MONA
in all three settings.
Together, these comparisons demonstrate the value
of constructing an additional spectral correction
through the parameter subproblem, complementing
acceleration through momentum design.

\paragraph{Fewer steps to baseline performance.}
Table~\ref{tab:step_reduction} quantifies the
optimization-step savings at matched validation loss.
AIMuon reaches Muon's final loss with
8.98\%, 10.05\%, and 6.49\% fewer steps
on GPT-2-small, GPT-2-medium, and Qwen3-0.6B,
respectively.
It also reaches the final losses of Muon-Nesterov
with 10.89\%, 4.13\%, and 3.78\% fewer steps.
Thus, the improvement extends beyond a lower
terminal loss: AIMuon reaches both baseline targets
earlier along the training trajectory.
Since methods within each setting process the same
number of tokens per step, these reductions also
correspond to fewer training tokens consumed
before reaching the respective targets.

\begin{table}[t]
    \centering
    \caption{
    Optimization-step reductions of AIMuon when reaching
    the final validation loss of each baseline.
    GPT-2-small reductions are computed from validation-loss
    curves averaged across three seeds;
    GPT-2-medium and Qwen3-0.6B use seed 42.
    }
    \label{tab:step_reduction}
    \small
    \begin{tabular}{lcc}
        \toprule
        Model & vs.\ Muon & vs.\ Muon-Nesterov \\
        \midrule
        GPT-2-small  & 8.98\%  & 10.89\% \\
        GPT-2-medium & 10.05\% & 4.13\%  \\
        Qwen3-0.6B   & 6.49\%  & 3.78\%  \\
        \bottomrule
    \end{tabular}
\end{table}

\subsection{Computational Cost}
\label{sec:wallclock}

At the same token budgets, GPT-2-medium training takes
4.50 hours with AIMuon and 4.26 hours with Muon,
corresponding to approximately 5.6\% additional training time.
For Qwen3-0.6B, the corresponding times are
14.96 and 14.36 hours, representing approximately
4.2\% additional training time.
All timings exclude validation.
Wall-clock performance curves and further discussion
are provided in Appendix~\ref{app:wallclock}.

\subsection{Component Analysis}
\label{sec:ablation}

We examine the contributions of the residual correction,
its independent spectral transformation,
and its relative weighting.
The no-residual variant tests the benefit of
the additional correction.
Inside-MSign and Linear residual examine its
placement and transformation, while the fixed-weight
variant examines adaptive nuclear-norm weighting.

All full-budget ablations use GPT-2-small,
2.48B tokens, and seed 42.
Inside-MSign and Linear residual are independently tuned
for learning rate and correction coefficient,
with $\beta=0.9$.
The no-residual variant is AIMuon with $c=0$.
Its retained Frobenius-norm matching introduces
numerical differences from the standalone Muon baseline.
The ablation update definitions are provided
in Appendix~\ref{app:ablation_configs}.

Figure~\ref{fig:aimuon_ablation} and
Table~\ref{tab:aimuon_ablation} summarize the results.
Full AIMuon achieves the lowest validation loss
among the evaluated configurations.

\begin{figure}[t]
    \centering
    \begin{minipage}[t]{0.49\linewidth}
        \vspace{0pt}
        \centering
        \includegraphics[width=\linewidth]
        {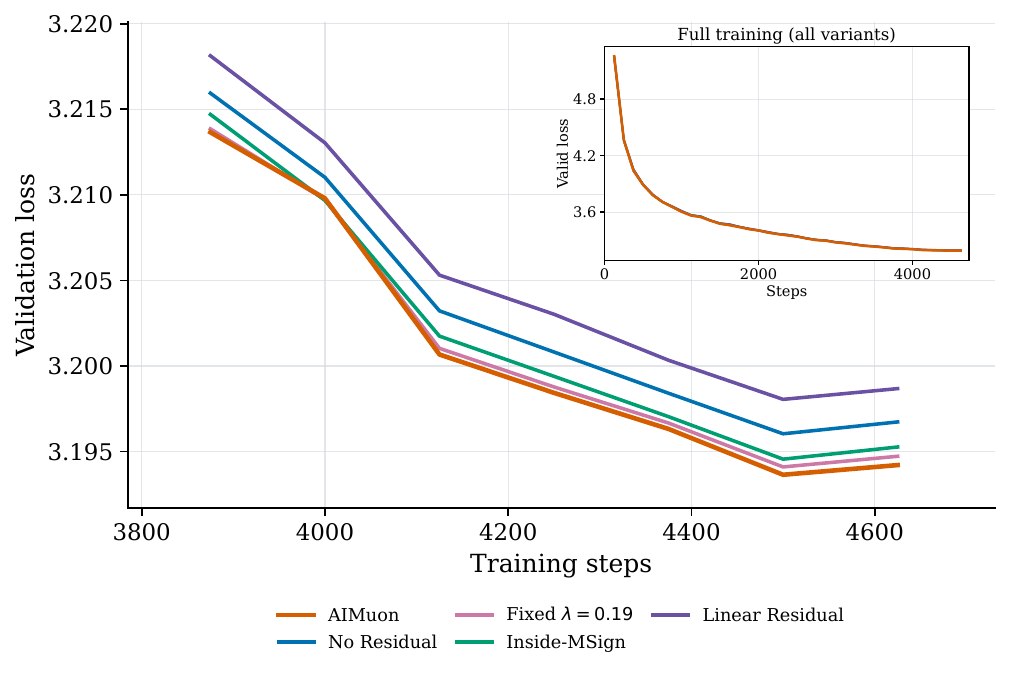}
        \caption{
GPT-2-small ablations using seed 42.
The main panel shows the final 20\% of training,
and the inset shows the full training trajectories.
}
        \label{fig:aimuon_ablation}
    \end{minipage}
    \hfill
    \begin{minipage}[t]{0.49\linewidth}
        \vspace{0pt}
        \centering
        \captionof{table}{
        Final validation loss for GPT-2-small ablations
        with 2.48B tokens and seed 42.
        }
        \label{tab:aimuon_ablation}
        \footnotesize
        \setlength{\tabcolsep}{3pt}
        \renewcommand{\arraystretch}{1.15}
        \begin{tabular}{@{}lcc@{}}
            \toprule
            Configuration
            & Coefficient
            & Loss $\downarrow$ \\
            \midrule
            AIMuon
            & $c=0.045$
            & \textbf{3.1939} \\
            Fixed residual weight
            & $\lambda_{\mathrm{fix}}=0.19$
            & 3.1944 \\
            Inside-MSign
            & $c=0.01$
            & 3.1949 \\
            No residual
            & $c=0$
            & 3.1964 \\
            Linear residual
            & $c=0.025$
            & 3.1984 \\
            \bottomrule
        \end{tabular}
    \end{minipage}
\end{figure}

\paragraph{Effectiveness of the residual correction.}
Removing the residual correction while retaining
AIMuon's normalization gives a final validation
loss of 3.1964.
Full AIMuon lowers this loss to 3.1939.
Both variants match their update magnitude
to the corresponding Muon reference, allowing
the comparison to assess the benefit of the
additional spectral residual direction
under the same magnitude-matching rule.

\paragraph{Why transform the residual separately?}
Inside-MSign and Linear residual distinguish
the effect of introducing residual information
from the effect of its independent spectral transformation.
Inside-MSign transforms the mixture $M_k+cE_k$,
whereas Linear residual adds $cE_k$ directly
to the transformed momentum.
After independent learning-rate and coefficient tuning,
these variants reach losses of 3.1949 and 3.1984,
respectively, compared with 3.1939 for AIMuon.

Inside-MSign improves upon the no-residual variant,
showing that incorporating the current gradient
through a momentum--residual mixture is useful.
AIMuon improves further by transforming the residual
separately before combining the two directions.
Meanwhile, adding the untransformed residual
does not improve upon the no-residual variant.
Together, these results support the effectiveness
of the residual's independent spectral transformation
beyond simply including residual information in the update.
This finding also complements the main comparison
with Muon-Nesterov, which mixes gradient and momentum
information before the spectral transformation.

\paragraph{Role of nuclear-norm weighting.}
Using a fixed residual weight of 0.19 gives
a validation loss of 3.1944, retaining most
of the improvement over the no-residual variant.
This result indicates that the separately transformed
residual direction accounts for most of the gain
in this comparison.
Adaptive nuclear-norm weighting further lowers
the loss to 3.1939 by adjusting the relative
contribution of the momentum and residual directions
across matrices and throughout training.
The full construction therefore combines
an effective additional direction with
a subproblem-derived rule for weighting it.

Additional sensitivity and batch-size analyses appear
in Appendix~\ref{app:correction_sensitivity}.
Neighboring correction coefficients yield similar
losses, showing that the improvement extends
across a range of correction strengths.
The batch-size experiments further characterize
how the benefit of the residual correction changes
with the training configuration, with a narrower
advantage at larger batch sizes.

\section{Conclusion}

We introduced AIMuon, which accelerates spectral optimization
through parameter-update design, complementing approaches
that modify momentum construction.
By further approximating the objective-containing parameter
subproblem in AIM, we jointly derive the gradient--momentum
residual, its independent spectral transformation,
and its nuclear-norm weight.
The relative nuclear-norm weights shape the combined
direction even after its magnitude is matched to Muon.
We establish a stochastic stationarity bound for the
unnormalized iteration with exact matrix-sign transformations.
Experiments on GPT-2-small, GPT-2-medium, and Qwen3-0.6B
show lower final validation loss than independently tuned
Muon, Muon-Nesterov, and MONA, with step reductions of
$8.98\%$, $10.05\%$, and $6.49\%$ relative to Muon's
final validation loss.
These results identify parameter-subproblem approximation
as a productive design direction for spectral optimizers.

\paragraph{Limitations and outlook.}
\label{sec:limitations}
AIMuon's additional spectral transformation increases
per-step computation, motivating more efficient
implementations and broader evaluation beyond
language-model pretraining.
The gradient--momentum residual contains both gradient
changes and stochastic fluctuations.
Understanding their interaction with spectral transformation
and residual weighting could guide adaptation across
batch sizes and noise levels.

AIMuon also motivates a closer examination of the
orthogonal structure of the update matrix.
Muon constructs an approximately orthogonal update,
whereas AIMuon combines two separately approximately
orthogonalized matrices whose weighted sum generally
no longer has a polar structure.
This provides a setting for examining how much of the
optimization benefit comes from the orthogonal structure
of the final update, and how much comes from reshaping
and combining the spectral structure of its components.

\clearpage

\section*{AI Use Statement}

Generative AI tools were used solely as auxiliary tools for language polishing and assistance with experimental and plotting code. The authors reviewed the resulting text and code and take full responsibility for the manuscript.

\bibliography{iclr2027_conference}
\bibliographystyle{iclr2027_conference}

\clearpage

\appendix
\section{Solution Set of the Spectral-Norm Subproblem}
\label{app:msign_solution_set}

We characterize the complete solution set of the spectral-norm penalized subproblem in \eqref{eq:prelim_msign_penalty}. The matrix-sign update used in the main text is then obtained by selecting a canonical representative from this set.

\begin{proposition}[Solution set of the spectral-norm subproblem]
\label{prop:msign_solution_set}
Let $A\in\mathbb{R}^{m\times n}$ and $\rho>0$, and consider
\begin{equation}
\min_Z\;
\left\{
\langle A,Z\rangle
+
\frac{\rho}{2}\|Z\|_2^2
\right\}.
\label{eq:app_msign_problem}
\end{equation}

If $A=0$, the unique minimizer is $Z^\star=0$. Otherwise, let $r=\operatorname{rank}(A)$ and write the compact SVD as $A=U_r\Sigma_rV_r^\top$. Let $U_\perp$ and $V_\perp$ denote orthonormal bases for the corresponding orthogonal complements. Then the optimal value is
\begin{equation}
-\frac{\|A\|_*^2}{2\rho},
\label{eq:app_msign_minimum_value}
\end{equation}
and the complete set of minimizers is
\begin{equation}
\begin{aligned}
&\operatorname*{arg\,min}_{Z}
\left\{
\langle A,Z\rangle
+
\frac{\rho}{2}\|Z\|_2^2
\right\}
\\
&\qquad=
\left\{
\frac{\|A\|_*}{\rho}
\left(
-U_rV_r^\top
+
U_\perp QV_\perp^\top
\right)
\;\middle|\;
Q\in\mathbb{R}^{(m-r)\times(n-r)},
\ \|Q\|_2\le1
\right\}.
\end{aligned}
\label{eq:app_msign_solution_set}
\end{equation}
\end{proposition}

\begin{proof}
The case $A=0$ is immediate, since the objective reduces to $\rho\|Z\|_2^2/2$. We therefore assume $A\neq0$. Let
\[
A=U_r\Sigma_rV_r^\top,
\qquad
\Sigma_r=\operatorname{diag}(\sigma_1,\ldots,\sigma_r),
\]
with $\sigma_i>0$, and set $a:=\|A\|_*>0$.

For any $Z$, let $t=\|Z\|_2$. By spectral--nuclear duality,
\begin{equation}
\langle A,Z\rangle
\ge
-\|A\|_*\|Z\|_2
=
-at.
\end{equation}
Hence
\begin{equation}
\begin{aligned}
\langle A,Z\rangle
+
\frac{\rho}{2}\|Z\|_2^2
&\ge
-at+\frac{\rho}{2}t^2\\
&=
\frac{\rho}{2}
\left(
t-\frac{a}{\rho}
\right)^2
-
\frac{a^2}{2\rho}\\
&\ge
-\frac{a^2}{2\rho}.
\end{aligned}
\label{eq:app_msign_lower_bound}
\end{equation}
The lower bound is attained by $Z=-(a/\rho)U_rV_r^\top$, since $\|U_rV_r^\top\|_2=1$ and $\langle A,U_rV_r^\top\rangle=a$. This proves \eqref{eq:app_msign_minimum_value}.

It remains to characterize all points attaining this bound. Equality in \eqref{eq:app_msign_lower_bound} requires
\begin{equation}
\|Z\|_2=\frac{a}{\rho},
\qquad
\langle A,Z\rangle=-\frac{a^2}{\rho}.
\label{eq:app_msign_optimality_equalities}
\end{equation}
Define
\[
C:=\frac{\rho}{a}Z.
\]
Then every minimizer satisfies
\begin{equation}
\|C\|_2=1,
\qquad
\langle A,C\rangle=-a.
\label{eq:app_msign_C_conditions}
\end{equation}

Let $u_i$ and $v_i$ denote the singular vectors associated with $\sigma_i$. From \eqref{eq:app_msign_C_conditions},
\[
\sum_{i=1}^r
\sigma_i u_i^\top Cv_i
=
-\sum_{i=1}^r\sigma_i.
\]
Since $\|C\|_2=1$, each $u_i^\top Cv_i\ge-1$. Because every $\sigma_i$ is positive, equality of the weighted sums forces
\[
u_i^\top Cv_i=-1,
\qquad i=1,\ldots,r.
\]
Moreover, $\|Cv_i\|_2\le1$, so
\[
\|Cv_i+u_i\|_2^2
=
\|Cv_i\|_2^2
+
2u_i^\top Cv_i
+
1
\le0.
\]
Therefore $Cv_i=-u_i$ for every $i$. Applying the same argument to $C^\top$ gives $C^\top u_i=-v_i$. Equivalently,
\begin{equation}
CV_r=-U_r,
\qquad
C^\top U_r=-V_r.
\label{eq:app_msign_singular_subspaces}
\end{equation}

Extend $U_r$ and $V_r$ to orthogonal matrices $U=[U_r\;U_\perp]$ and $V=[V_r\;V_\perp]$. The relations in \eqref{eq:app_msign_singular_subspaces} imply that $C$ has the block form
\begin{equation}
U^\top C V
=
\begin{pmatrix}
-I_r & 0\\
0 & Q
\end{pmatrix},
\qquad
Q=U_\perp^\top C V_\perp.
\label{eq:app_msign_block_form}
\end{equation}
Since $\|C\|_2=1$, necessarily $\|Q\|_2\le1$. Transforming back gives
\begin{equation}
C
=
-U_rV_r^\top
+
U_\perp QV_\perp^\top,
\end{equation}
and hence
\begin{equation}
Z
=
\frac{\|A\|_*}{\rho}
\left(
-U_rV_r^\top
+
U_\perp QV_\perp^\top
\right),
\qquad
\|Q\|_2\le1.
\end{equation}

Conversely, any matrix of this form is optimal. Indeed, for any $Q$ with $\|Q\|_2\le1$, the block representation \eqref{eq:app_msign_block_form} gives
\[
\|C\|_2
=
\max\{1,\|Q\|_2\}
=
1,
\]
while the orthogonality of the complementary subspaces gives $\langle A,C\rangle=-a$. Thus \eqref{eq:app_msign_C_conditions} holds, so \eqref{eq:app_msign_optimality_equalities} holds as well. Therefore every such $Z$ attains the optimal value in \eqref{eq:app_msign_minimum_value}, completing the characterization in \eqref{eq:app_msign_solution_set}.
\end{proof}

Choosing $Q=0$ in \eqref{eq:app_msign_solution_set} gives
\begin{equation}
Z_{\mathrm{can}}
=
-\frac{\|A\|_*}{\rho}U_rV_r^\top
=
-\frac{\|A\|_*}{\rho}\operatorname{MSign}(A),
\end{equation}
which is the canonical representative used in \eqref{eq:prelim_msign_penalty_solution}. If $r=\min(m,n)$, one of the complementary subspaces is trivial, so the additional term vanishes and the minimizer is unique. If $0<r<\min(m,n)$, different admissible choices of $Q$ yield distinct minimizers.

\section{Stochastic Convergence Analysis}
\label{app:aimuon_stochastic_fro}

We analyze the exact AIMuon iteration, excluding Newton--Schulz approximation errors, Frobenius-norm matching, gradient clipping, weight decay, and auxiliary AdamW updates. For a matrix $A\in\mathbb{R}^{m\times n}$, define
\begin{equation}
\mathcal{D}(A)
:=
\|A\|_*\operatorname{MSign}(A),
\qquad
\mathcal{D}(0):=0,
\end{equation}
where $\operatorname{MSign}(A)=U_qV_q^\top$ for the compact SVD of $A$ containing its $q=\operatorname{rank}(A)$ positive singular values. We consider the exact iteration
\begin{equation}
\begin{aligned}
M_k
&=
\beta M_{k-1}+(1-\beta)G_k,\\
E_k
&=
G_k-M_k,\\
\mathcal{U}_k
&=
\mathcal{D}(M_k)+c\mathcal{D}(E_k),\\
\Theta_{k+1}
&=
\Theta_k-\eta_k\mathcal{U}_k,
\qquad M_{-1}=0.
\end{aligned}
\label{eq:aimuon_exact_update_fro}
\end{equation}
We take $\Theta_0$ to be deterministic. For the single-matrix iteration considered here, a fixed positive scaling factor can be absorbed into $\eta_k$.

\subsection{Assumptions and Main Result}

\begin{assumption}[Lower boundedness and smoothness]
\label{ass:aimuon_smooth_fro}
The objective $F:\mathbb{R}^{m\times n}\to\mathbb{R}$ is continuously differentiable, bounded below by $F_{\inf}>-\infty$, and has an $L_F$-Lipschitz continuous gradient for some $L_F>0$:
\begin{equation}
\|\nabla F(X)-\nabla F(Y)\|_F
\le
L_F\|X-Y\|_F,
\qquad
\forall X,Y.
\label{eq:aimuon_fro_smooth_assumption}
\end{equation}
\end{assumption}

\begin{assumption}[Unbiased stochastic gradients]
\label{ass:aimuon_noise_fro}
Let $\mathcal{F}_k$ denote the filtration generated by the history prior to sampling $G_k$, so that $\Theta_k$ and $M_{k-1}$ are $\mathcal{F}_k$-measurable. For deterministic batch sizes $B_k\ge1$, assume
\begin{equation}
\begin{aligned}
\mathbb{E}[G_k\mid\mathcal{F}_k]
&=
\nabla F(\Theta_k),\\
\mathbb{E}\!\left[
\|G_k-\nabla F(\Theta_k)\|_F^2
\,\middle|\,
\mathcal{F}_k
\right]
&\le
\frac{\sigma_F^2}{B_k}.
\end{aligned}
\label{eq:aimuon_fro_variance_assumption}
\end{equation}
\end{assumption}

\begin{theorem}[Stochastic Frobenius-stationarity bound]
\label{thm:aimuon_stochastic_fro}
Suppose Assumptions~\ref{ass:aimuon_smooth_fro} and~\ref{ass:aimuon_noise_fro} hold. Let $0\le\beta<1$ and $0<c<4$. Define
\begin{equation}
r:=\min(m,n),
\qquad
\kappa_c:=\frac{4-c}{4(1+2c)},
\qquad
\Delta_F:=F(\Theta_0)-F_{\inf}.
\end{equation}
Assume that the stepsizes $\{\eta_k\}$ are deterministic, nonnegative, and nonincreasing, with
\begin{equation}
0<\eta_0
\le
\frac{\kappa_c}{rL_F},
\qquad
S_T
:=
\sum_{k=0}^{T-1}\eta_k
>0.
\label{eq:aimuon_fro_stepsizes}
\end{equation}
Then the iterates generated by \eqref{eq:aimuon_exact_update_fro} satisfy
\begin{equation}
\begin{aligned}
\frac{1}{S_T}
\sum_{k=0}^{T-1}
\eta_k
\mathbb{E}
\|\nabla F(\Theta_k)\|_F^2
&\le
\frac{4\Delta_F}{c\kappa_c S_T}\\
&\quad+
\frac{4r\sigma_F^2}{c\kappa_c^2 S_T}
\sum_{k=0}^{T-1}
\frac{\eta_k}{B_k}.
\end{aligned}
\label{eq:aimuon_fro_stochastic_bound}
\end{equation}
In particular, for constant stepsize $\eta_k\equiv\eta>0$ and constant batch size $B_k\equiv B$,
\begin{equation}
\frac{1}{T}
\sum_{k=0}^{T-1}
\mathbb{E}
\|\nabla F(\Theta_k)\|_F^2
\le
\frac{4\Delta_F}{c\kappa_c\eta T}
+
\frac{4r\sigma_F^2}{c\kappa_c^2 B}.
\label{eq:aimuon_fro_constant_batch}
\end{equation}
\end{theorem}

For constant stepsize and batch size, \eqref{eq:aimuon_fro_constant_batch} consists of an $O(T^{-1})$ optimization term and an $O(B^{-1})$ stochastic noise floor.

\subsection{Preliminary Inequalities}

\begin{lemma}[Properties of the nuclear-weighted polar map]
\label{lem:aimuon_polar_fro}
Let $r=\min(m,n)$ and $\mathcal{H}(A)=\tfrac12\|A\|_*^2$. The map $\mathcal{D}$ satisfies
\begin{equation}
\begin{aligned}
\langle A,\mathcal{D}(A)\rangle
&=
\|A\|_*^2,\\
\|\mathcal{D}(A)\|_2
&=
\|A\|_*,\\
\|\mathcal{D}(A)\|_F^2
&=
\operatorname{rank}(A)\|A\|_*^2
\le
r\|A\|_*^2,\\
\mathcal{D}(A)
&\in
\partial\mathcal{H}(A).
\end{aligned}
\label{eq:aimuon_polar_fro_properties}
\end{equation}
All identities remain valid at $A=0$.
\end{lemma}

\begin{proof}
For $A\ne0$, write its compact SVD as $A=U_q\Sigma_qV_q^\top$. Then
\begin{equation}
\langle A,U_qV_q^\top\rangle
=
\|A\|_*,
\qquad
\|U_qV_q^\top\|_2
=
1,
\qquad
\|U_qV_q^\top\|_F^2
=
q.
\end{equation}
This proves the norm and inner-product identities. For any matrix $B$, spectral--nuclear duality gives
\begin{equation}
\langle\mathcal{D}(A),B\rangle
\le
\|A\|_*\|B\|_*
\le
\tfrac12\|A\|_*^2+\tfrac12\|B\|_*^2.
\end{equation}
Rearranging and using $\langle A,\mathcal{D}(A)\rangle=\|A\|_*^2$ yields
\begin{equation}
\mathcal{H}(B)
\ge
\mathcal{H}(A)
+
\langle\mathcal{D}(A),B-A\rangle.
\end{equation}
This is the subgradient inequality. The zero case follows from $\mathcal{D}(0)=0$ and $\mathcal{H}(B)\ge0$.
\end{proof}

\begin{lemma}[Coercive quadratic bounds]
\label{lem:aimuon_quadratic_fro}
For $0<c<4$, let
\[
\kappa_c
=
\frac{4-c}{4(1+2c)}
\]
and
\[
q_c(a,b)
=
a^2-cab+cb^2.
\]
For all $a,b\ge0$,
\begin{equation}
q_c(a,b)
\ge
c\kappa_c(a+b)^2,
\qquad
q_c(a,b)
\ge
\kappa_c(a+cb)^2.
\label{eq:aimuon_quadratic_fro_bounds}
\end{equation}
\end{lemma}

\begin{proof}
Both inequalities follow from completing the square:
\begin{equation}
\begin{aligned}
q_c(a,b)-c\kappa_c(a+b)^2
&=
\frac{((c+2)a-3cb)^2}{4(1+2c)},\\
q_c(a,b)-\kappa_c(a+cb)^2
&=
\frac{c(3a-(c+2)b)^2}{4(1+2c)}.
\end{aligned}
\end{equation}
\end{proof}

\subsection{Proof of Theorem~\ref{thm:aimuon_stochastic_fro}}

\begin{proof}
All inner products below are Frobenius inner products. Nuclear and spectral norms appear only as proof tools for the unchanged exact update.

All expectations below are finite at each finite iteration. Indeed, Lemma~\ref{lem:aimuon_polar_fro} and $\|A\|_*\le\sqrt{r}\|A\|_F$ give
\[
\|\mathcal{D}(A)\|_F
\le
r\|A\|_F.
\]
The Lipschitz gradient has at most linear growth, and the conditional variance bound then gives finite second moments of $G_k$ whenever $\Theta_k$ has a finite second moment. Starting from fixed $\Theta_0$ and $M_{-1}=0$, the momentum and parameter recursions therefore imply finite second moments of all states by induction. Smoothness and lower boundedness also make $F(\Theta_k)$ integrable.

Write
\begin{equation}
\begin{aligned}
H_k
&:=
\nabla F(\Theta_k),
&
\zeta_k
&:=
G_k-H_k,
&
\tau
&:=
\frac{\beta}{1-\beta},
\\
a_k
&:=
\|M_k\|_*,
&
b_k
&:=
\|E_k\|_*,
&
Q_k
&:=
a_k^2-ca_kb_k+cb_k^2.
\end{aligned}
\end{equation}

The momentum recursion implies the exact identity
\begin{equation}
E_k
=
\tau(M_k-M_{k-1}).
\label{eq:aimuon_momentum_identity_fro}
\end{equation}
Since $\mathcal{D}(M_k)\in\partial\mathcal{H}(M_k)$,
\begin{equation}
\begin{aligned}
\langle E_k,\mathcal{D}(M_k)\rangle
&=
\tau
\langle M_k-M_{k-1},\mathcal{D}(M_k)\rangle\\
&\ge
\tau
\bigl(
\mathcal{H}(M_k)-\mathcal{H}(M_{k-1})
\bigr).
\end{aligned}
\label{eq:aimuon_momentum_energy_fro}
\end{equation}

Using $G_k=M_k+E_k$ and $\langle M_k,\mathcal{D}(E_k)\rangle\ge-a_kb_k$, we obtain
\begin{equation}
\begin{aligned}
\langle G_k,\mathcal{U}_k\rangle
&=
a_k^2
+
cb_k^2
+
\langle E_k,\mathcal{D}(M_k)\rangle
+
c\langle M_k,\mathcal{D}(E_k)\rangle\\
&\ge
Q_k
+
\tau
\bigl(
\mathcal{H}(M_k)-\mathcal{H}(M_{k-1})
\bigr).
\end{aligned}
\label{eq:aimuon_alignment_energy_fro}
\end{equation}

Lemmas~\ref{lem:aimuon_polar_fro} and~\ref{lem:aimuon_quadratic_fro} give
\begin{equation}
\|\mathcal{U}_k\|_F^2
\le
r(a_k+cb_k)^2
\le
\frac{r}{\kappa_c}Q_k.
\label{eq:aimuon_update_size_fro}
\end{equation}
Also,
\[
\|G_k\|_F
\le
\|G_k\|_*
\le
a_k+b_k,
\]
hence
\begin{equation}
Q_k
\ge
c\kappa_c\|G_k\|_F^2.
\label{eq:aimuon_sampled_gradient_fro}
\end{equation}

Set $\eta_{-1}=\eta_0$ and define the nonnegative energy
\begin{equation}
\mathcal{V}_k
:=
F(\Theta_k)-F_{\inf}
+
\tau\eta_{k-1}\mathcal{H}(M_{k-1}).
\label{eq:aimuon_lyapunov_fro}
\end{equation}
Frobenius smoothness and $H_k=G_k-\zeta_k$ imply
\begin{equation}
\begin{aligned}
F(\Theta_{k+1})
&\le
F(\Theta_k)
-
\eta_k\langle H_k,\mathcal{U}_k\rangle
+
\frac{L_F\eta_k^2}{2}
\|\mathcal{U}_k\|_F^2\\
&\le
F(\Theta_k)
-
\tau\eta_k
\bigl(
\mathcal{H}(M_k)-\mathcal{H}(M_{k-1})
\bigr)\\
&\quad
-
\eta_k
\left(
1-\frac{rL_F\eta_k}{2\kappa_c}
\right)
Q_k
+
\eta_k
\langle\zeta_k,\mathcal{U}_k\rangle.
\end{aligned}
\end{equation}

Because $\eta_k\le\kappa_c/(rL_F)$ and $\eta_k\le\eta_{k-1}$, adding the new momentum energy gives
\begin{equation}
\mathcal{V}_{k+1}
\le
\mathcal{V}_k
-
\frac{\eta_k}{2}Q_k
+
\eta_k
\langle\zeta_k,\mathcal{U}_k\rangle.
\end{equation}

By Frobenius Cauchy--Schwarz and Young's inequality,
\begin{equation}
\begin{aligned}
\langle\zeta_k,\mathcal{U}_k\rangle
&\le
\|\zeta_k\|_F
\sqrt{\frac{rQ_k}{\kappa_c}}\\
&\le
\frac14Q_k
+
\frac{r}{\kappa_c}\|\zeta_k\|_F^2.
\end{aligned}
\end{equation}
Therefore,
\begin{equation}
\mathcal{V}_{k+1}
\le
\mathcal{V}_k
-
\frac{\eta_k}{4}Q_k
+
\frac{r\eta_k}{\kappa_c}
\|\zeta_k\|_F^2.
\label{eq:aimuon_stochastic_energy_fro}
\end{equation}

Unbiasedness in the Frobenius inner-product space gives
\begin{equation}
\mathbb{E}
\left[
\|G_k\|_F^2
\mid
\mathcal{F}_k
\right]
=
\|H_k\|_F^2
+
\mathbb{E}
\left[
\|\zeta_k\|_F^2
\mid
\mathcal{F}_k
\right].
\end{equation}
Combining this identity with \eqref{eq:aimuon_sampled_gradient_fro}, we have
\begin{equation}
\mathbb{E}
\left[
Q_k
\mid
\mathcal{F}_k
\right]
\ge
c\kappa_c\|H_k\|_F^2.
\end{equation}

Taking conditional expectations in \eqref{eq:aimuon_stochastic_energy_fro} yields
\begin{equation}
\begin{aligned}
\mathbb{E}
\left[
\mathcal{V}_{k+1}
\mid
\mathcal{F}_k
\right]
&\le
\mathcal{V}_k
-
\frac{c\kappa_c\eta_k}{4}
\|H_k\|_F^2
+
\frac{r\eta_k\sigma_F^2}
{\kappa_c B_k}.
\end{aligned}
\end{equation}

Taking total expectations and summing over $k=0,\ldots,T-1$, and using $M_{-1}=0$, $\mathcal{V}_0=\Delta_F$, and $\mathcal{V}_T\ge0$, gives
\begin{equation}
\frac{c\kappa_c}{4}
\sum_{k=0}^{T-1}
\eta_k
\mathbb{E}
\|\nabla F(\Theta_k)\|_F^2
\le
\Delta_F
+
\frac{r\sigma_F^2}{\kappa_c}
\sum_{k=0}^{T-1}
\frac{\eta_k}{B_k}.
\end{equation}
Dividing by $c\kappa_c S_T/4$ proves \eqref{eq:aimuon_fro_stochastic_bound}. Setting $\eta_k\equiv\eta$ and $B_k\equiv B$ gives \eqref{eq:aimuon_fro_constant_batch}.

Importantly, $\mathcal{U}_k$ depends nonlinearly on $G_k$. The proof therefore does not set
\[
\mathbb{E}
\left[
\langle\zeta_k,\mathcal{U}_k\rangle
\mid
\mathcal{F}_k
\right]
\]
to zero; instead, this term is bounded explicitly before taking expectations.
\end{proof}

\section{Additional Experimental Details}
\label{app:experiment_details}

\subsection{Training Configuration}

\paragraph{Models, datasets, and budgets.}
We pretrain GPT-2-small on FineWeb-Edu and
GPT-2-medium and Qwen3-0.6B on FineWeb.
All models are initialized from scratch.
Table~\ref{tab:training_configuration} summarizes
the main training configurations.
All three settings process 524,288 tokens per
optimization step, with training budgets rounded
up to an integer number of steps.

\begin{table}[t]
    \centering
    \caption{
    Main pretraining configurations.
    Batch size is the global number of sequences per step.
    }
    \label{tab:training_configuration}
    \small
    \begin{tabular}{lcccc}
        \toprule
        Model & Dataset & Tokens & Sequence length & Batch size \\
        \midrule
        GPT-2-small  & FineWeb-Edu & 2.48B & 1024 & 512 \\
        GPT-2-medium & FineWeb     & 6.9B  & 1024 & 512 \\
        Qwen3-0.6B   & FineWeb     & 12B   & 4096 & 128 \\
        \bottomrule
    \end{tabular}
\end{table}

\paragraph{Data processing.}
Documents are tokenized using the corresponding
model's tokenizer, separated by EOS tokens,
and deterministically packed into contiguous sequences.
The first 4,096 packed sequences form a fixed
validation set and are excluded from training.
Within each setting and seed, methods share
initialization, training-data order, batch configuration,
and validation data.

\paragraph{Numerical settings and hardware.}
All experiments use BF16 mixed precision,
global gradient clipping with a maximum norm of 1.0,
and distributed data-parallel training on eight
NVIDIA H100 GPUs.

\paragraph{Learning-rate schedule.}
Both search runs and full-budget runs use linear
warmup over the first 1\% of optimization steps,
followed by cosine decay to zero over the remaining 99\%.
Each run uses a complete schedule scaled to its
own training budget.
Full-budget runs are evaluated every 125 steps
and at the final step.

\subsection{Optimizer Implementation}

\paragraph{Parameter grouping.}
The matrix optimizer is applied to eligible
two-dimensional weight matrices in the Transformer
hidden layers.
Fused query, key, and value projections, where present,
are treated as a single matrix.
Embeddings, normalization parameters, biases,
the language-modeling head, and remaining parameters
use AdamW.
The AdamW fallback uses the same base learning-rate
schedule as the matrix optimizer, without
matrix-specific dimension scaling.

\paragraph{Momentum and matrix-sign inputs.}
The ordinary EMA momentum and gradient--momentum
residual are
\[
    M_k=\beta M_{k-1}+(1-\beta)G_k,
    \qquad M_{-1}=0,
    \qquad E_k=G_k-M_k.
\]
Muon applies the matrix-sign approximation directly
to $M_k$.
Muon-Nesterov retains the same momentum state
but transforms
\[
    N_k=\beta M_k+(1-\beta)G_k.
\]
AIMuon separately transforms momentum and residual:
\[
    D_k^M=\operatorname{NS}_5(M_k),
    \qquad
    D_k^E=\operatorname{NS}_5(E_k).
\]

\paragraph{Newton--Schulz approximation.}
All matrix-sign transformations use five
Newton--Schulz iterations.
The input is cast to BF16 and transposed when
its number of rows exceeds its number of columns.
It is normalized by the larger of its Frobenius
norm and $\epsilon_{\mathrm{NS}}=10^{-7}$.
We apply five iterations with polynomial coefficients
$(3.4445,-4.7750,2.0315)$ and transpose the output
back when necessary.

\paragraph{Nuclear-norm weighting and magnitude matching.}
AIMuon estimates the nuclear norms using
\[
    \widehat{\nu}_k^M
    =
    |\langle M_k,D_k^M\rangle|,
    \qquad
    \widehat{\nu}_k^E
    =
    |\langle E_k,D_k^E\rangle|.
\]
The relative residual weight and combined direction are
\[
    \lambda_k
    =
    c\frac{\widehat{\nu}_k^E}
    {\widehat{\nu}_k^M+\varepsilon},
    \qquad
    \widetilde{O}_k
    =
    \widehat{\nu}_k^M
    \left(D_k^M+\lambda_kD_k^E\right).
\]
The final direction is
\[
    O_k
    =
    0.2\sqrt{\max(m,n)}
    \frac{\|D_k^M\|_F}
    {\|\widetilde{O}_k\|_F+\varepsilon}
    \widetilde{O}_k,
    \qquad \varepsilon=10^{-12}.
\]
Matching the combined magnitude to the corresponding
Muon update controls the scale change introduced
by the residual correction, allowing us to assess
the benefit of modifying the update direction.

\paragraph{Other optimizer parameters.}
All matrix optimizers use weight decay of 0.1.
For MONA, we use $\beta_a=0.99$ and
$\alpha=-1/[2(1-\beta_a)]=-50$.
The AdamW fallback uses
$(\beta_1,\beta_2)=(0.9,0.95)$ and
$\epsilon_{\mathrm{Adam}}=10^{-8}$.
Its decay-group weight decay is 0.01 for GPT-2
and 0.1 for Qwen3-0.6B.
Normalization parameters, biases, and other
no-decay parameters use zero weight decay.
AIMuon uses $c=0.045$, $0.0475$, and $0.05$
for GPT-2-small, GPT-2-medium, and Qwen3-0.6B,
respectively.

\subsection{Hyperparameter Selection}

\paragraph{Search protocol.}
We independently tune the learning rate and momentum
coefficient of AIMuon, Muon, Muon-Nesterov, and MONA.
For each optimizer, we first search the peak learning
rate at a fixed initial momentum coefficient.
We then fix the selected learning rate and search
$\beta\in\{0.9,0.95\}$.
Other hyperparameters are fixed within each search stage.

Every candidate is trained from scratch using seed 42,
with matched initialization, training-data order,
and validation data.
Selection uses terminal validation loss.
The selected configurations are used for full-budget
training with freshly initialized model and optimizer states.
Table~\ref{tab:main_search_configuration} summarizes
the learning-rate grids and per-candidate budgets.

\begin{table}[t]
    \centering
    \caption{
    Main-optimizer search configurations.
    Learning-rate candidates are listed in units of
    $10^{-3}$.
    The token budget applies to each candidate in
    both the learning-rate and momentum searches.
    }
    \label{tab:main_search_configuration}
    \small
    \setlength{\tabcolsep}{4pt}
    \begin{tabular}{llcc}
        \toprule
        Model
        & Learning-rate candidates
        & Initial $\beta$
        & Tokens/candidate \\
        \midrule
        GPT-2-small
        & $\{2,2.5,3,3.5,4,4.5,5\}$
        & 0.95 & 2.48B \\
        GPT-2-medium
        & $\{1,2,3,4,4.5,5\}$
        & 0.95 & 0.69B \\
        Qwen3-0.6B
        & $\{1,2,3,3.5,4,4.5\}$
        & 0.9 & 1.2B \\
        \bottomrule
    \end{tabular}
\end{table}

\paragraph{Selected configurations.}
All four optimizers independently select a peak
learning rate of $4.5\times10^{-3}$ for both GPT-2
settings and $4.0\times10^{-3}$ for Qwen3-0.6B.
The momentum search selects $\beta=0.9$ for all
four optimizers in both GPT-2 settings.
For Qwen3-0.6B, it selects $\beta=0.95$ for AIMuon,
Muon, and Muon-Nesterov, and $\beta=0.9$ for MONA.

Each GPT-2-small search candidate uses the full
2.48B-token budget, corresponding to 4,731 steps.
For GPT-2-medium and Qwen3-0.6B, each candidate
uses 10\% of the corresponding full training budget.

\subsection{Ablation Configurations}
\label{app:ablation_configs}

\paragraph{Common configuration.}
All full-budget ablations use GPT-2-small on
FineWeb-Edu with 2.48B training tokens, seed 42,
sequence length 1024, and global batch size 512.
Initialization, training-data order, and validation
data are matched across variants.
All reported configurations use a peak learning rate
of $4.5\times10^{-3}$ and momentum coefficient
$\beta=0.9$.

\paragraph{Update definitions.}
All variants use the EMA momentum
$M_k=\beta M_{k-1}+(1-\beta)G_k$
and gradient--momentum residual $E_k=G_k-M_k$.
The residual-construction controls use the
unnormalized directions
\begin{equation}
\begin{aligned}
\text{Fixed residual weight:}\quad
D_k
&=
\operatorname{NS}_5(M_k)
+
\lambda_{\mathrm{fix}}\operatorname{NS}_5(E_k),
\\
\text{Inside-MSign:}\quad
D_k
&=
\operatorname{NS}_5(M_k+cE_k),
\\
\text{Linear residual:}\quad
D_k
&=
\operatorname{NS}_5(M_k)+cE_k.
\end{aligned}
\label{eq:ablation_directions}
\end{equation}
The fixed-weight variant replaces AIMuon's
nuclear-norm-dependent relative weight with
a constant $\lambda_{\mathrm{fix}}$.
Inside-MSign combines momentum and residual before
applying a single spectral transformation.
Linear residual adds the untransformed residual
directly to the spectral momentum direction.

The no-residual variant is AIMuon with $c=0$,
retaining its nuclear-norm scaling and final
Frobenius-norm matching.
Its unnormalized direction is therefore
\[
    D_k
    =
    \widehat{\nu}_k^M\operatorname{NS}_5(M_k).
\]
The retained Frobenius-norm matching introduces
numerical differences from the standalone Muon baseline.

\paragraph{Shared magnitude matching.}
For each parameter matrix in $\mathbb{R}^{m\times n}$,
the directions are rescaled using
\begin{equation}
    O_k
    =
    0.2\sqrt{\max(m,n)}
    \frac{\|\operatorname{NS}_5(M_k)\|_F}
    {\|D_k\|_F+\varepsilon}
    D_k.
    \label{eq:ablation_magnitude_matching}
\end{equation}
The parameter update is
\begin{equation}
    \Theta_{k+1}
    =
    (1-\eta_k\lambda_{wd})\Theta_k-\eta_kO_k.
    \label{eq:ablation_parameter_update}
\end{equation}
This rule matches the update magnitude to the
corresponding Muon reference, allowing the comparisons
to distinguish the benefit of changing the update
direction from that of increasing the update magnitude.

\paragraph{Fixed residual weight.}
We set $\lambda_{\mathrm{fix}}=0.19$ based on
the logged mean residual weight
(\texttt{lambda\_mean}) around the midpoint
of the reference AIMuon training run.
This provides a constant-weight comparison
at a representative correction strength.

\paragraph{Hyperparameter selection for residual controls.}
Inside-MSign and Linear residual are independently
tuned using the same candidate grids:
\[
    c\in\{0.01,\,0.025,\,0.05,\,0.1,\,0.2\},
    \qquad
    \eta_{\max}\in\{4.0,\,4.5,\,5.0\}\times10^{-3}.
\]
In both the correction-coefficient and learning-rate
searches, each candidate uses 0.248B tokens,
seed 42, and a complete warmup and cosine-decay
schedule for its allotted budget.
The momentum coefficient is fixed at $\beta=0.9$,
and selection is based on terminal validation loss.

Inside-MSign initially selects the lower endpoint,
$c=0.01$, so we additionally evaluate $c=0.005$
under the same protocol; $c=0.01$ remains better.
Linear residual selects $c=0.025$.
Both variants select a peak learning rate of
$4.5\times10^{-3}$.
Each selected configuration is then restarted
from scratch and trained for the full 2.48B-token
budget using seed 42.

The full AIMuon reference uses $c=0.045$.
The no-residual variant uses $c=0$ and the
independently selected Muon learning rate and
momentum coefficient reported above.
Its final validation loss is 3.1964, and the same
result is used for the $c=0$ entry in the
residual-coefficient sensitivity analysis.

\subsection{Evaluation Protocol}

\paragraph{Final validation performance.}
GPT-2-small results report the mean and standard
deviation across seeds 42, 43, and 44.
Hyperparameters are selected using seed 42
and retained for seeds 43 and 44.
GPT-2-medium and Qwen3-0.6B use seed 42.
All comparisons use matched seeds.

\paragraph{Steps to baseline performance.}
We use the final validation losses of Muon
and Muon-Nesterov as separate targets.
For GPT-2-small, we first average validation losses
across seeds 42, 43, and 44 at each evaluation step.
We then interpolate the mean validation-loss curves
to determine when AIMuon first reaches each
baseline's final mean validation loss.
For GPT-2-medium and Qwen3-0.6B, we apply the same
procedure to the seed-42 curves.

For baseline $b$, the step reduction is
\[
    R_b
    =
    \frac{T_b-T_{\mathrm{AIMuon}\to b}}{T_b}
    \times100\%,
\]
where $T_b$ is the baseline's full training length
and $T_{\mathrm{AIMuon}\to b}$ is the interpolated
step at which AIMuon first reaches the target.
All crossings are measured along the original
full-budget learning-rate schedules.

\paragraph{Wall-clock measurements.}
Wall-clock measurements exclude validation time
and include all training computation, including
AIMuon's additional residual transformation.

\subsection{Sensitivity and Batch-Size Experiments}
\label{app:correction_sensitivity}

\paragraph{Residual-coefficient sensitivity.}
We evaluate
\[
    c\in\{0,\,0.025,\,0.045,\,0.075,\,0.1\}
\]
on GPT-2-small.
All configurations use 2.48B training tokens,
seed 42, global batch size 512,
peak learning rate $4.5\times10^{-3}$, and $\beta=0.9$.
Each configuration with $c>0$ is trained from scratch.
The $c=0$ entry reuses the no-residual AIMuon
ablation, retaining its Frobenius-norm matching,
with a final validation loss of 3.1964.

\begin{figure}[t]
    \centering
    \includegraphics[width=0.85\linewidth]
    {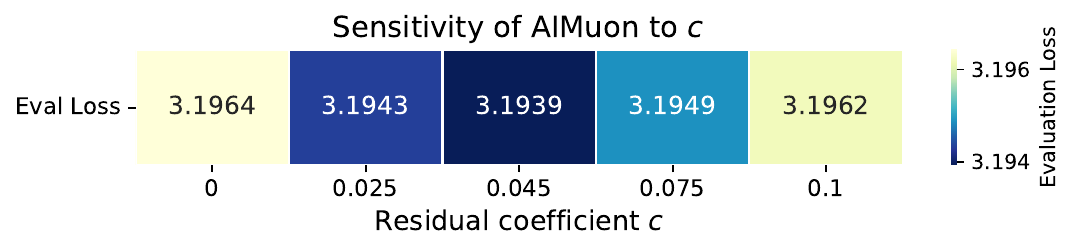}
    \caption{
    Sensitivity to the residual coefficient on GPT-2-small
    with 2.48B tokens and seed 42.
    The $c=0$ entry uses AIMuon without residual correction,
    retaining its Frobenius-norm matching.
    Lower validation loss is better.
    }
    \label{fig:residual_coefficient_sensitivity}
\end{figure}

Figure~\ref{fig:residual_coefficient_sensitivity}
shows the final validation losses.
All tested nonzero coefficients improve upon
the no-residual variant,
with the lowest loss at $c=0.045$.
The neighboring values $c=0.025$ and $c=0.075$
yield similar losses, showing that the improvement
extends across a range of correction strengths.
Increasing the coefficient does not monotonically
improve performance.

\paragraph{Batch-size dependence.}
We compare AIMuon, Muon, and Muon-Nesterov
on GPT-2-small with global batch sizes
\[
    B\in\{128,\,256,\,512,\,1024,\,2048\}.
\]
All runs use a sequence length of 1024,
approximately 2.48B tokens, seed 42,
peak learning rate $4.5\times10^{-3}$,
and $\beta=0.9$.
Each run uses a complete learning-rate schedule
scaled to its number of optimization steps.
The AIMuon correction coefficients are
\[
\begin{array}{c|ccccc}
    B & 128 & 256 & 512 & 1024 & 2048 \\
    \hline
    c & 0.05 & 0.05 & 0.045 & 0.025 & 0.01
\end{array}
\]
The token budget is fixed across batch sizes,
so larger batches use fewer optimizer updates.

\begin{figure}[t]
    \centering
    \includegraphics[width=\linewidth]
    {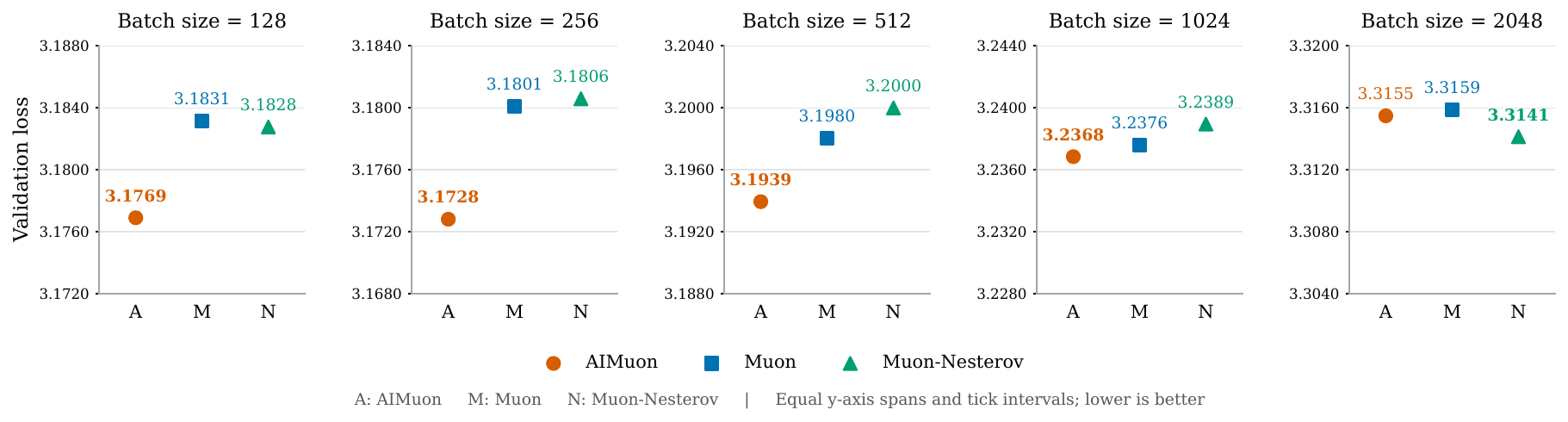}
    \caption{
    Final validation loss across global batch sizes
    on GPT-2-small with 2.48B tokens and seed 42.
    A, M, and N denote AIMuon, Muon, and Muon-Nesterov,
    respectively.
    }
    \label{fig:batch_size_comparison}
\end{figure}

Figure~\ref{fig:batch_size_comparison} shows
that AIMuon achieves the lowest final validation
loss at batch sizes 128--1024, while Muon-Nesterov
performs best at 2048.
AIMuon's advantage over Muon narrows at larger
batch sizes.
These results characterize how the benefit of
residual correction varies with batch size
and correction strength under a fixed token budget.

\subsection{Wall-Clock Performance}
\label{app:wallclock}

AIMuon performs an additional Newton--Schulz
transformation for the residual, increasing
its per-step computational cost.
The full-budget timings reported in
Section~\ref{sec:wallclock} measure this overhead
at a fixed token budget.
The time required to reach a target validation loss
depends on both the per-step cost and the number
of optimization steps.

\begin{figure}[t]
    \centering
    \begin{subfigure}[t]{0.32\linewidth}
        \centering
        \includegraphics[width=\linewidth]
        {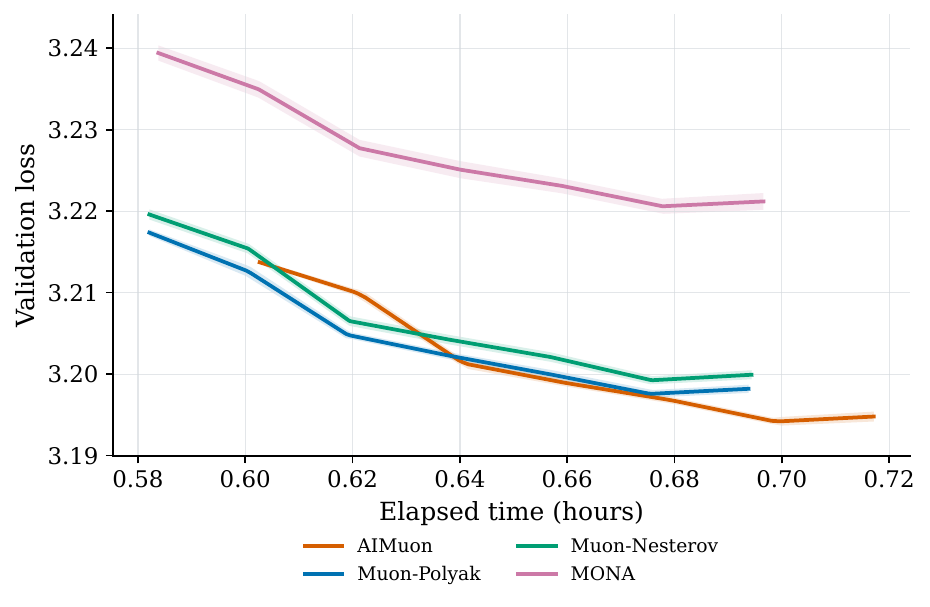}
        \caption{GPT-2-small.}
        \label{fig:small_wallclock}
    \end{subfigure}
    \hfill
    \begin{subfigure}[t]{0.32\linewidth}
        \centering
        \includegraphics[width=\linewidth]
        {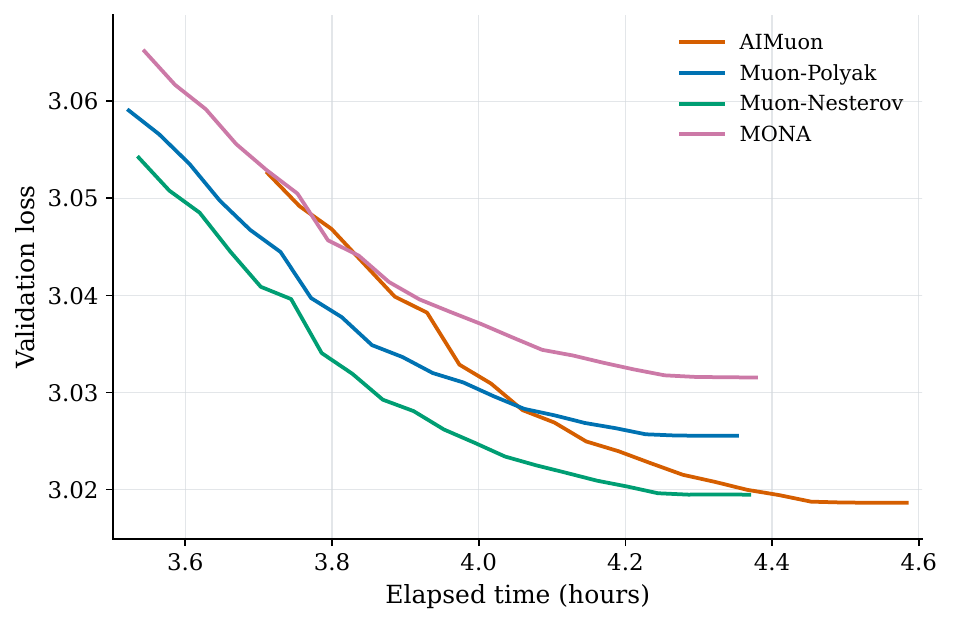}
        \caption{GPT-2-medium.}
        \label{fig:medium_wallclock}
    \end{subfigure}
    \hfill
    \begin{subfigure}[t]{0.32\linewidth}
        \centering
        \includegraphics[width=\linewidth]
        {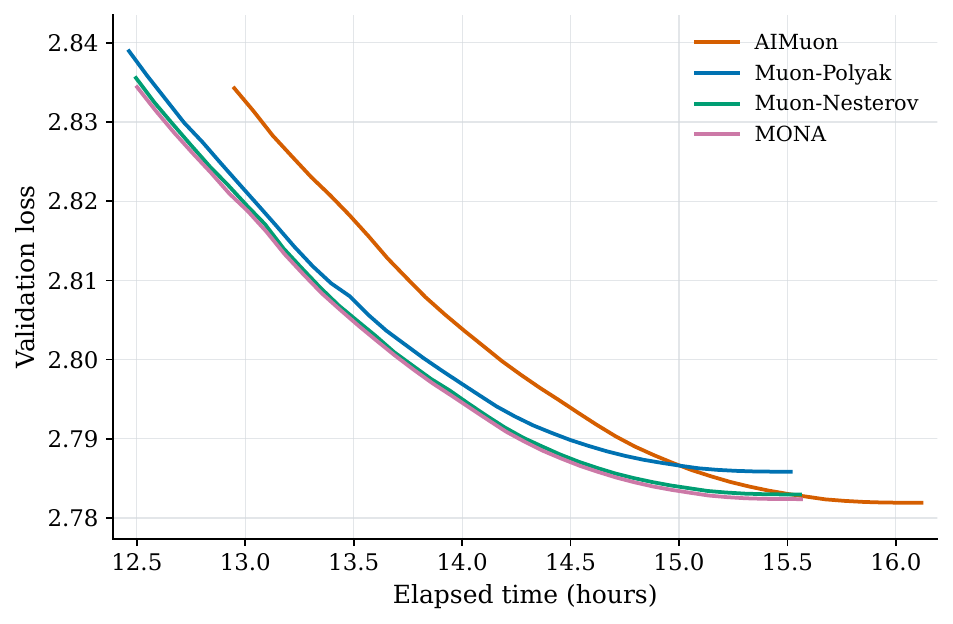}
        \caption{Qwen3-0.6B.}
        \label{fig:qwen_wallclock}
    \end{subfigure}
    \caption{
    Validation loss versus elapsed training time,
    excluding validation time, during the final portion
    of pretraining on eight NVIDIA H100 GPUs.
    All trajectories use their original
    full-budget learning-rate schedules.
    }
    \label{fig:wallclock_comparison}
\end{figure}

Figure~\ref{fig:wallclock_comparison} compares
validation loss against elapsed training time
on GPT-2-small, GPT-2-medium, and Qwen3-0.6B.
On GPT-2-small, AIMuon attains lower loss than
the baselines within the later part of the
overlapping time range.
On Qwen3-0.6B, Muon-Nesterov and MONA achieve
lower losses over most of the displayed overlapping
time range, while AIMuon reaches a lower terminal
loss after additional training time.
These curves show the combined effect of
per-step computational cost and optimization progress.

\clearpage
\end{CJK*}
\end{document}